\documentclass[letterpaper, 10 pt, conference]{ieeeconf}  

\IEEEoverridecommandlockouts                              

\usepackage{amsmath,amssymb,amsfonts}
\usepackage{algorithmic}
\usepackage{graphicx}
\usepackage{textcomp}
\usepackage{xcolor}
\usepackage{tikz}
\usepackage{soul}
\sethlcolor{yellow}
\usepackage{booktabs}
\usetikzlibrary{fit,backgrounds}
\usetikzlibrary{arrows.meta,positioning,calc}
\usetikzlibrary{positioning,fit,backgrounds,calc}
\usepackage{cite}
\usepackage[colorlinks=true,citecolor=blue,linkcolor=blue,urlcolor=blue]{hyperref}
\usepackage{pifont}
\usepackage{xcolor}
\definecolor{venuegray}{gray}{0.45}

\def\BibTeX{{\rm B\kern-.05em{\sc i\kern-.025em b}\kern-.08em
    T\kern-.1667em\lower.7ex\hbox{E}\kern-.125emX}}

\title{\LARGE \bf
Bend the Clock: Predicting Ahead to Beat Latency in Event-Based Object Detection
}

\author{
Biswadeep Sen$^{1,2}$, Benoit R. Cottereau$^{2,3}$, Nicolas Cuperlier$^{2,4}$, Terence Sim$^{1,2}$ \\\\
\\
$^{1}$ National University of Singapore, Singapore \\
$^{2}$ IPAL CNRS IRL 2955, Singapore \\
$^{3}$ CerCo, CNRS UMR 5549, Universit\'{e} de Toulouse, France \\
$^{4}$ ETIS UMR8051, CY Cergy Paris Universit\'{e}, ENSEA, CNRS, Cergy, France
}

\begin{document}

\maketitle
\thispagestyle{empty}
\pagestyle{empty}

\begin{abstract}
Event cameras promise low-latency perception for high-speed robotic systems,
where even short delays can render detections stale by the time they inform
downstream robotic decisions. Yet modern event detectors still require tens
of milliseconds of computation before their predictions become available.
Conventional evaluation ignores this delay by comparing predictions with
annotations at the observation timestamp, even though the scene may have
changed by the time those predictions are produced. We study this
\emph{observation--availability mismatch} in event-based multi-object
detection and show that state-of-the-art event detectors degrade substantially
when evaluated at prediction availability rather than observation time. To
address this, we introduce \textbf{ChronoFuse}, a causal availability-time
detector that predicts object states for when its output becomes available
rather than for when its input was observed. ChronoFuse performs causal
cross-time fusion over a multi-scale feature hierarchy, combining current
representations with cached temporal features to expose short-term temporal
cues without using future observations. The fusion pathway is lightweight,
adding only \(0.17\) million parameters and \(0.84\,\mathrm{ms}\) of mean
end-to-end latency overhead. ChronoFuse recovers \(71\%\) of the accuracy lost
to latency on 1Mpx driving data and \(90.8\%\) under rapid drone motion on
FRED, nearly restoring zero-delay performance. Under the extreme motion of
EV-Flying, ChronoFuse reaches \(20.95\) sAP, compared with \(2.25\) for the
strongest standard event detector (\(9.3\times\)). These
results show that predicting ahead can be critical for robots operating in
fast-changing scenes, including autonomous driving, agile flight, and robotic
interception.
\end{abstract}

\section{Introduction}

Event cameras promise perception at the speed of motion. By reporting
brightness changes asynchronously rather than capturing complete frames,
they offer microsecond-scale temporal resolution, high dynamic range, and
little motion blur~\cite{gallego2022event}. These properties make them
attractive for high-speed robotic perception, including autonomous driving,
agile flight, and robotic interception, where a few milliseconds can
separate a useful detection from an obsolete one. Yet low sensing latency
does not guarantee timely perception. Events must still be accumulated,
encoded, processed, transferred, and post-processed. A detector observing
the world at time \(t\) therefore returns its prediction only at
\(t+\Delta\). By then, an accurate box may already describe the past.

\begin{figure*}[t]
    \centering
    \includegraphics[width=\textwidth]{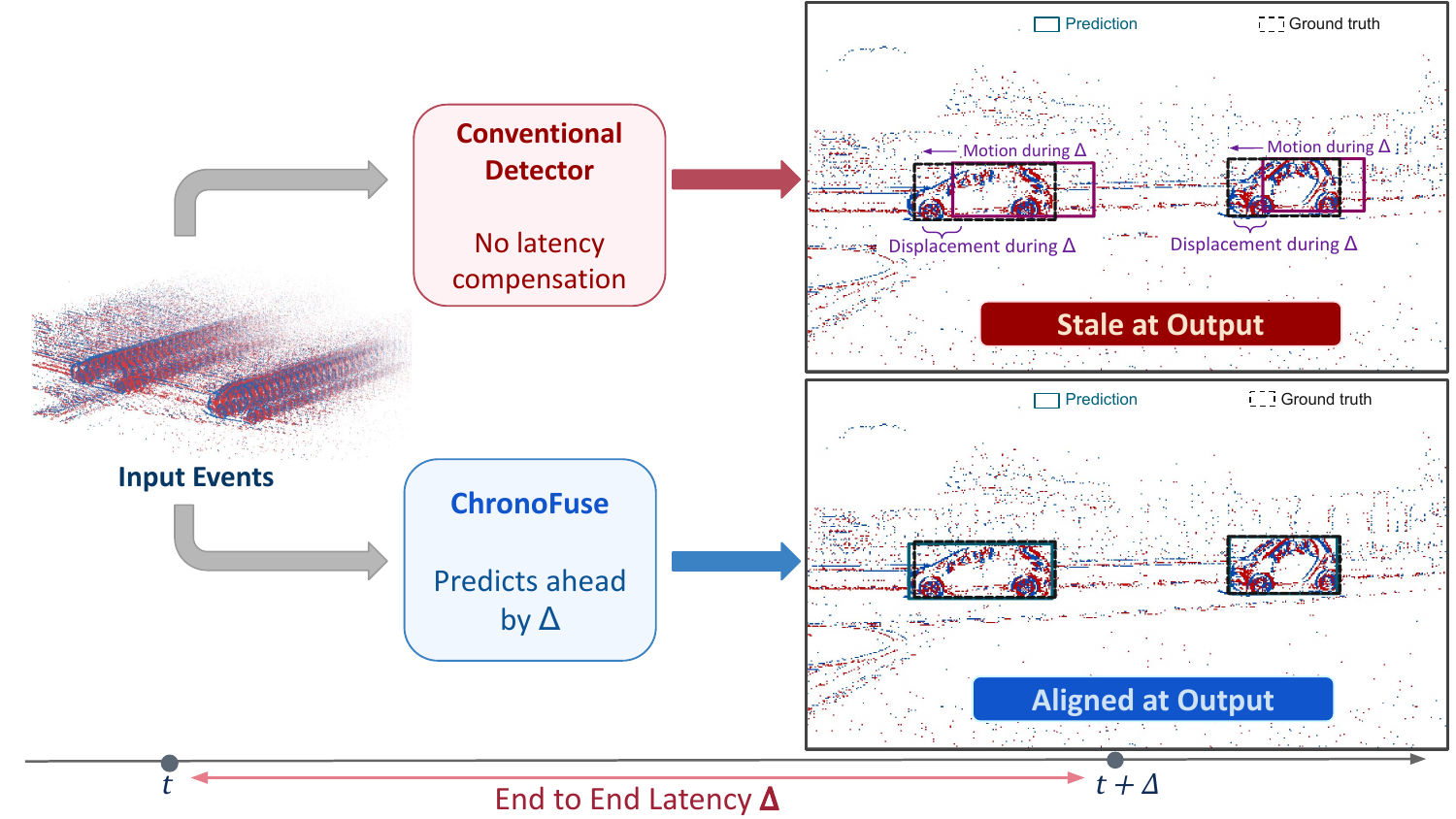}
    \caption{\textbf{Computation turns accurate detections stale.}
    An event window ending at \(t\) is processed with end-to-end latency
    \(\Delta\). Conventional detection becomes stale as objects move during
    computation, while ChronoFuse predicts ahead by \(\Delta\) to align
    detections with the scene at output time.}
    \label{fig:latency_motivation}
\end{figure*}

Conventional detection evaluation conceals this failure mode. Offline AP
compares predictions derived from observations at \(t\) with annotations at
the same timestamp, effectively granting the detector zero computation time.
In a deployed robot, however, downstream planning and control receive those
predictions only at \(t+\Delta\), after objects may have moved, appeared, or
left the scene. Stale localization can therefore propagate into robotic
decisions even when the detector was accurate for its original input. As
illustrated in Fig.~\ref{fig:latency_motivation}, this delay leaves a
conventional detector spatially stale at output time, whereas predicting
ahead by \(\Delta\) aligns the detection with the state of the world
available to the robot. Despite their low-latency sensors, modern event
detectors still require tens of milliseconds end-to-end
(Table~\ref{tab:e2e_latency}). After only \(33.3\,\mathrm{ms}\), even perfect
boxes from time \(t\) retain just \(72.96\) mAP on 1Mpx~\cite{gen4}. The
central question is therefore not only \emph{how accurately can we detect the
present?}, but \emph{what will be present when the prediction becomes
available?}

This mismatch is not unique to event vision. In the far more mature RGB
detection literature, streaming perception exposed the gap between offline
evaluation and real-time deployment, motivating detectors that predict ahead
to compensate inference delay~\cite{li2020towards,yang2022real,
zhang2025transtreaming}. However, these methods are developed for
frame-domain inputs; they neither characterize how availability-time error
manifests in event detectors nor exploit the recurrent representations
formed from asynchronous events. Existing event detectors use temporal
memory to improve detection, but are still trained and evaluated to describe
the latest observation rather than the scene at prediction availability.

This gap is especially consequential in event vision. The rapid motion that
makes event sensing most valuable also causes observation-time detections to
become stale most quickly. Although faster models reduce the delay, they do
not change the timestamp the detector is trained to describe. Exploiting the
temporal advantage of event cameras therefore requires not only faster
detection, but predicting the state that will be available to downstream
robotic planning and control.

We address this gap with \textbf{ChronoFuse}, which formulates event detection
around the time at which a prediction becomes available:
\[
    \hat{\mathcal{Y}}_{t+\Delta}
    =
    f_{\theta}\!\left(\mathcal{E}_{\leq t}\right).
\]

Here, \(\mathcal{E}_{\leq t}\) denotes the event history observed through time
\(t\), and \(\hat{\mathcal{Y}}_{t+\Delta}\) the predicted object classes and
bounding boxes at the future availability time. ChronoFuse combines
future-target supervision with a causal \textbf{Cross-Time Fusion Module}
that integrates current and cached multi-scale features, providing the
detection head with ordered short-term temporal cues without using future
observations. The fused representation is decoded directly into
availability-time detections rather than extrapolating already-decoded
boxes. The fusion pathway remains lightweight, adding only 0.17M parameters
and increasing mean end-to-end latency by \(0.84\,\mathrm{ms}\).

The benefit is already substantial on 1Mpx driving data, where ChronoFuse
recovers \(71\%\) of the accuracy lost to latency. The gains become striking
for rapid aerial motion: on FRED, ChronoFuse recovers \(90.8\%\) of the
latency-induced loss, nearly restoring same-time detection performance.
EV-Flying exposes the failure mode at its extreme: observation-time event
detectors reach at most $2.25$ sAP under streaming evaluation, whereas
ChronoFuse reaches $20.95$ sAP---$9.3\times$ the best standard
detector. These results show that predictive compensation becomes
increasingly important for robotic perception precisely in the high-speed
regimes where event cameras are most valuable.

Our contributions are threefold:
\begin{itemize}

\item We quantify the \emph{observation--availability mismatch} in
event-based detection, showing that observation-time evaluation
overstates the accuracy delivered to a robot.

\item We introduce \textbf{ChronoFuse}, combining future-target
supervision with causal cross-time fusion to align detections with
prediction availability.

\item We demonstrate its effectiveness across driving, drone, and
bird/insect detection, recovering $90.8\%$ of the latency-induced loss
on FRED and achieving $9.3\times$ the sAP of the best observation-time
detector on EV-Flying.

\end{itemize}
\section{Related Work}

\subsection{Event-Based Object Detection}

Event-based object detection has progressed from convolutional recurrence to
increasingly efficient Transformer and state-space architectures. Early
recurrent models established temporal memory for event streams
~\cite{gen4}, while RVT combined recurrence with hierarchical
spatial attention to achieve a strong accuracy--runtime trade-off
~\cite{rvt}. Subsequent methods targeted the cost of processing
long event sequences more aggressively: SAST sparsified attention over
informative windows and tokens~\cite{sast}, S5-ViT replaced recurrent
aggregation with continuous-time state-space dynamics~\cite{ssm},
and SMamba combined sparse event representations with selective state-space
modeling~\cite{smamba}. More recently, SSLA-Det further reduced
computation through spatially sparse linear attention, though its evaluation
was limited to the smaller-scale Gen1 and N-Caltech101 benchmarks
~\cite{hao2026lowlatency}. Collectively, these methods have pushed both
accuracy and efficiency substantially forward, making event-based detection
increasingly attractive for real-time robotic perception.

Despite these advances, event detectors remain fundamentally
observation-time predictors: improving runtime reduces staleness but does not
predict through the computation delay. This distinction becomes increasingly
important as object motion grows, since even a short delay can displace a box
substantially before it reaches the downstream system. Among related event
detectors, RED included one-step-ahead regression only as auxiliary
regularization; its actual detection output remained observation-time aligned
and therefore did not provide latency-compensated streaming detection
~\cite{gen4}. ChronoFuse instead performs latency-compensated streaming
detection by making availability-time prediction the primary detection
objective.

\subsection{Streaming and Latency-Aware Perception}

Frame-based streaming perception formalized the gap between observation time
and prediction availability. Li et al.\ introduced streaming AP (sAP), which
scores the latest available prediction against the scene at the evaluation
time rather than against the frame that originally produced it
~\cite{li2020towards}. StreamYOLO showed that inference delay could be
compensated by predicting future detections rather than only accelerating the
detector~\cite{yang2022real}. LongShortNet and DAMO-StreamNet extended this
idea through short- and long-term temporal feature fusion
~\cite{li2023longshortnet,he2023damo}, while Transtreaming further extended
prediction across multiple future horizons to accommodate varying inference
delay~\cite{zhang2025transtreaming}. Together, these works established
predictive compensation for RGB video, but were built around regularly
sampled frames and frame-centric temporal fusion. Event-only detectors operate
on event-derived temporal representations, making direct transfer of these
streaming designs a poor fit.

Event-based work has largely pursued latency reduction rather than prediction
through the computation delay. SODFormer used asynchronous event--frame fusion
~\cite{li2023sodformer}, while DAGr reduced repeated computation through
incremental event updates~\cite{gehrig2024lowlatency}. Both shortened the
observation-to-output path, but their detections still described the latest
observed scene; finite processing therefore can leave fast-moving objects
vulnerable to staleness. STARE studied latency-aware evaluation and prediction
in single-object tracking, where a target was initialized from a supplied box
rather than discovered and classified by a detector~\cite{chu2026stare}.

A complementary approach is to compensate motion after detection.
Constant-velocity and Kalman extrapolation are inexpensive ~\cite{kalman1960new,li2020towards}, but operate only on completed detections,
depend on temporal association and motion assumptions, and cannot recover
objects missed before extrapolation. Prior event-forecasting methods such as
E-Motion and E-TIDE predicted future event representations rather than object
detections~\cite{wu2024emotion,sen2026etide}. Converting these forecasts into
detections requires a second inference stage, adding computation and allowing
forecasting errors to propagate into the final boxes. ChronoFuse instead
predicts object detections directly at availability time, without
reconstructing future events or extrapolating completed detections.

\section{Method}
\label{sec:method}

ChronoFuse treats detector latency as a temporal-alignment problem. At
processing step $k$, the latest causal observation ends at $t_k$, but
the resulting detections become usable only after computation has
finished. A conventional detector describes the scene at $t_k$, so its
output may already be stale when consumed. ChronoFuse addresses this
through two complementary components: future-target supervision shifts
the desired output to $t_k+\Delta$, while a Cross-Time Fusion Module
combines the preceding and current feature pyramids to capture recent
scene evolution. Both components use only events observed no later than
$t_k$.

\begin{figure*}[t]
    \centering
    \includegraphics[width=\textwidth]{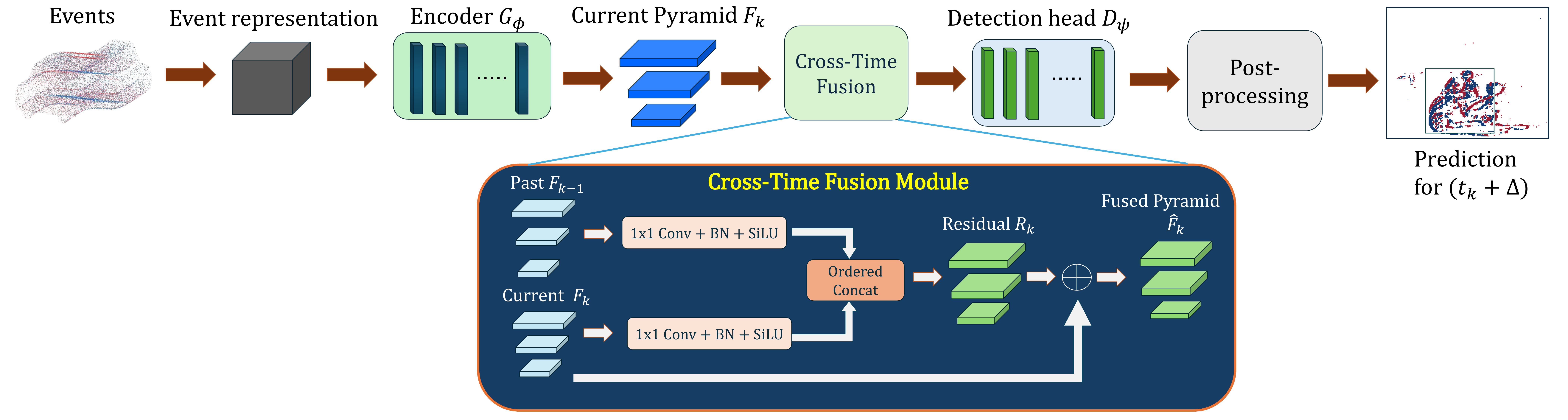}
    \caption{
    \textbf{Overview of ChronoFuse.} Events observed up to the causal cutoff
    $t_k$ are encoded into the current feature pyramid $\mathcal{F}_k$.
    The Cross-Time Fusion Module combines it with the cached raw pyramid
    $\mathcal{F}_{k-1}$ to produce the fused representation
    $\widetilde{\mathcal{F}}_k$, which is passed to the YOLOX head.
    During training, the raw prediction $Z_k$ is supervised using the
    matched future target $\mathcal{Y}_k^{+}$; post-processing is used
    only to obtain the final prediction for $t_k+\Delta$.
    }
    \label{fig:chronofuse_architecture}
\end{figure*}

The complete architecture and the internal structure of the Cross-Time
Fusion Module are shown in Fig.~\ref{fig:chronofuse_architecture}. A
causal event representation is passed through a recurrent encoder to
obtain the current feature pyramid. The module then combines each
pyramid level with its counterpart cached from the preceding step. The
resulting fused pyramid is passed to the detection head. Training pairs
each causal input with boxes and classes near $t_k+\Delta$; removing the
Cross-Time Fusion Module gives the future-target-only (FT-only) variant.

\subsection{Latency-Compensated Detection}
\label{sec:method_problem}

An event is represented as $e_i=(x_i,y_i,\tau_i,p_i)$, where
$(x_i,y_i)$ is its pixel location, $\tau_i$ its timestamp, and
$p_i\in\{-1,+1\}$ its polarity. At each processing cutoff $t_k$, only
events observed during the preceding integration window are available.
Let $\mathcal R$ denote the event-representation operator that converts
these events into the tensor consumed by the detector. The causal input
is therefore
\begin{equation}
X_k
=
\mathcal R\!\left(
\{e_i\mid
t_k-T_{\mathrm{win}}<\tau_i\leq t_k\}
\right),
\label{eq:event_repr}
\end{equation}
where $T_{\mathrm{win}}$ is the integration-window duration. By
construction, $X_k$ contains no event occurring after $t_k$.

The representation is next processed by a recurrent encoder and
multi-scale feature neck. Denote this network by $\mathcal G_\phi$,
where $\phi$ comprises its trainable parameters. At step $k$, it
produces the current feature pyramid $\mathcal F_k$ and updates the
recurrent state $S_k$:
\begin{equation}
\begin{aligned}
(\mathcal F_k,S_k)
&=\mathcal G_\phi(X_k,S_{k-1}),\\
\mathcal F_k
&=\{F_k^{(\ell)}\}_{\ell=1}^{L}.
\end{aligned}
\label{eq:causal_encoder}
\end{equation}
Here, $F_k^{(\ell)}\in
\mathbb R^{C_\ell\times H_\ell\times W_\ell}$ is the feature map at
pyramid level $\ell$. We instantiate $\mathcal G_\phi$ using the RVT
encoder and feature pyramid~\cite{rvt}.

Let $\mathcal Y_q(t)$ denote the ground-truth boxes and classes in
sequence $q$ at time $t$. An observation-time detector is trained to
estimate $\mathcal Y_q(t_k)$. ChronoFuse instead estimates the scene at
$t_k+\Delta$, while retaining exactly the same causal observation
support.

For completeness, let $\widetilde{\mathcal F}_k$ denote the feature
pyramid presented to the detection head. It equals $\mathcal F_k$ in
the FT-only variant and is the temporally fused pyramid in ChronoFuse,
as defined in Sec.~\ref{sec:method_fusion}. Let $\mathcal D_\psi$ denote
the YOLOX-style detection head with parameters $\psi$. Its output
$Z_k$ contains the raw box-regression, objectness, and classification
predictions. The operator $\operatorname{Post}$ subsequently performs
decoding, score filtering, and non-maximum suppression:
\begin{equation}
\begin{aligned}
Z_k
&=\mathcal D_\psi(\widetilde{\mathcal F}_k),\\
\widehat{\mathcal Y}_k^{(\Delta)}
&=\operatorname{Post}(Z_k)
\approx \mathcal Y_q(t_k+\Delta).
\end{aligned}
\label{eq:future_detection}
\end{equation}

The horizon $\Delta$ changes only the time represented by the desired
output. Events in $(t_k,t_k+\Delta]$ remain unavailable during both
training and inference. The integration window, interval between
processing cutoffs, prediction horizon, and measured runtime are
distinct temporal quantities. Their dataset-specific values are given
in Sec.~\ref{sec:experimental_setup}.

\subsection{Future-Target Supervision}
\label{sec:method_future}

Whereas conventional supervision uses the observed configuration
$\mathcal Y_q(t_k)$, we train ChronoFuse against the object configuration
one prediction horizon later:
\begin{equation}
\mathcal Y_k^+
=
\mathcal Y_q(t_k+\Delta).
\label{eq:future_target}
\end{equation}
Here, $\mathcal Y_k^+$ denotes the valid annotation temporally closest
to $t_k+\Delta$ within the same sequence. The common prediction horizon and its relation to measured
end-to-end detector latency are specified in
Sec.~\ref{sec:experimental_setup}.

\subsection{Cross-Time Fusion Module}
\label{sec:method_fusion}

The Cross-Time Fusion Module exposes recent scene evolution by
presenting the detection head with the preceding and current feature
pyramids. At pyramid level $\ell$, the temporal input is
\begin{equation}
H_k^{(\ell)}=
\begin{cases}
F_{k-1}^{(\ell)}, & \text{within a sequence},\\
F_k^{(\ell)},     & \text{at its first step}.
\end{cases}
\label{eq:feature_cache}
\end{equation}
Thus, $H_k^{(\ell)}$ contains only previously computed information,
while boundary initialization prevents cross-sequence leakage.

At each pyramid level, the cached and current features are processed by
the same projection. Specifically, for
$U\in\{H_k^{(\ell)},F_k^{(\ell)}\}$, we define
\begin{equation}
\begin{aligned}
P_\ell(U)
&=\operatorname{SiLU}\!\left(
  \operatorname{BN}_\ell\!\left(
  \operatorname{Conv}^{(\ell)}_{1\times1}(U)
  \right)\right),\\
P_\ell(U)
&\in\mathbb R^{
  \frac{C_\ell}{2}\times H_\ell\times W_\ell}.
\end{aligned}
\label{eq:temporal_projection}
\end{equation}
The parameters of $P_\ell$ are shared between the cached and current
branches but are independent across pyramid levels.

The Cross-Time Fusion Module concatenates the projected features in
fixed past--present order and adds the result to the current feature:
\begin{equation}
\begin{aligned}
R_k^{(\ell)}
&=\left[
P_\ell(H_k^{(\ell)})
\,\Vert\,
P_\ell(F_k^{(\ell)})
\right],\\
\widetilde F_k^{(\ell)}
&=F_k^{(\ell)}+R_k^{(\ell)},\\
\widetilde{\mathcal F}_k
&=\{\widetilde F_k^{(\ell)}\}_{\ell=1}^{L}.
\end{aligned}
\label{eq:chronofuse}
\end{equation}
Here, $\Vert$ denotes channel-wise concatenation. Each branch produces
$C_\ell/2$ channels, so $R_k^{(\ell)}$ has the same dimensions as
$F_k^{(\ell)}$. The fixed ordering preserves temporal direction, while
the residual path retains the complete current representation. The
Cross-Time Fusion Module does not explicitly estimate flow, velocity,
or correspondence; the correction is learned directly from
future-target supervision.

We zero-initialize the affine BatchNorm parameters:
\begin{equation}
\begin{aligned}
\gamma_\ell=\beta_\ell=0
&\Longrightarrow P_\ell(U)=0,\\
&\Longrightarrow R_k^{(\ell)}=0,\\
&\Longrightarrow
\widetilde F_k^{(\ell)}=F_k^{(\ell)}.
\end{aligned}
\label{eq:zero_init}
\end{equation}
The initial model therefore preserves the pretrained detection function,
after which optimization progressively learns the temporal residual.

\subsection{Optimization and Causal Execution}
\label{sec:method_training}

Let $\phi$, $\omega$, and $\psi$ denote the parameters of the feature
encoder, Cross-Time Fusion Module, and detection head, respectively, and
define $\Theta=(\phi,\omega,\psi)$. Let $\mathcal K_\Delta$ be the set of training steps having a valid
future target $\mathcal Y_k^+$. Training minimizes
\begin{equation}
\Theta^\star =
\arg\min_{\Theta}\,
\frac{1}{|\mathcal{K}_\Delta|}
\sum_{k\in\mathcal{K}_\Delta}
\mathcal{L}_{\mathrm{det}}(Z_k,\mathcal{Y}_k^+).
\label{eq:training_objective}
\end{equation}
For the YOLOX-style head~\cite{yolox}, SimOTA assignment is recomputed
using the future boxes; the original regression, objectness, and
classification losses remain unchanged. Steps outside
$\mathcal K_\Delta$ contribute no detection loss but still update the
temporal state and feature cache.

The FT-only ablation removes the Cross-Time Fusion Module while
retaining the same future targets, detector, and training objective,
i.e., $\widetilde{\mathcal F}_k=\mathcal F_k$. Its comparison with
ChronoFuse therefore isolates the contribution of explicit cross-time
feature fusion.

After prediction, only the unfused current pyramid is cached:
\begin{equation}
H_{k+1}^{(\ell)}\leftarrow F_k^{(\ell)}.
\label{eq:cache_update}
\end{equation}
Consequently, temporal residuals do not accumulate recursively.
ChronoFuse requires no additional encoder evaluation and stores only
one feature pyramid per stream. The cache and temporal state are reset
at sequence boundaries.
\section{Experiments}
\label{sec:experiments}

We evaluate whether predicting object locations ahead of the current
event timestamp can compensate for the accuracy degradation introduced
by inference latency. Our experiments address four questions:
(i) how strongly does latency degrade event-based object detection,
(ii) can learned future prediction recover the lost accuracy better than
simple motion extrapolation,
(iii) does the advantage persist under increasingly fast motion, and
(iv) which components of the proposed future-fusion architecture
contribute to its performance.

\subsection{Experimental Setup}
\label{sec:experimental_setup}

\paragraph{Datasets}
We evaluate on 1Mpx~\cite{gen4}, FRED~\cite{fred}, and
EV-Flying~\cite{evflying}, spanning increasingly challenging
latency-sensitive regimes. 1Mpx represents autonomous-driving
scenarios with cars and vulnerable road users such as pedestrians,
cyclists, and motorcyclists, where detection latency can directly induce
safety-critical localization errors. FRED shifts to aerial robotics, with
small, fast-moving drones exhibiting less constrained motion and requiring
timely localization for tracking and navigation. Finally, EV-Flying provides
a stringent stress test, with rapidly moving birds and insects whose small
size, abrupt trajectories, and large displacements make stale detections
particularly severe. This progression tests whether latency compensation
remains effective as target motion becomes less constrained and object
scale decreases.


\paragraph{Streaming metrics}
Following prior work on streaming perception~\cite{yang2022real,zhang2025transtreaming},
we report streaming Average Precision (sAP), which measures detection
accuracy at the time predictions become available rather than at the input
timestamp. We report COCO-style $\mathrm{sAP}_{50:95}$, together with
$\mathrm{sAP}_{50}$ and $\mathrm{sAP}_{75}$. All values are reported as
percentages.

\paragraph{Latency-aware streaming evaluation}
Let $E_{\leq t}$ denote all events observed up to time $t$, and let
$B_{\tau}$ denote the ground-truth detections at time $\tau$. For a detector
with end-to-end latency $\Delta$, a prediction computed from observations
available at $t$ reaches the downstream system only at $t+\Delta$.
Accordingly, streaming evaluation compares the prediction with the scene at
its availability time:
\begin{equation}
    \hat{B}_{t}=f_{\theta}(E_{\leq t}),
    \qquad
    \hat{B}_{t}\longleftrightarrow B_{t+\Delta}.
    \label{eq:stream_eval}
\end{equation}
Thus, even an accurate prediction for the observed scene may become stale
during computation. ChronoFuse instead predicts the target state at
$t+\Delta$ using only events available up to $t$, and is evaluated against
the same ground truth $B_{t+\Delta}$.

Unless otherwise stated, we evaluate at a common streaming horizon of
$\Delta=33.33$\,ms (one 30\,Hz time step), which lies within the measured
end-to-end latency range of contemporary event-based detectors
(Table~\ref{tab:e2e_latency}). For RVT, this falls between its mean
($29.89$\,ms) and P90 ($37.27$\,ms) latency. All timings are measured on an
NVIDIA RTX~3090 and include representation construction, inference,
prediction transfer, decoding, and NMS. Input construction remains dataset-specific: 1Mpx uses
$50\,\mathrm{ms}$ event-integration windows and processing strides,
whereas FRED and EV-Flying use $33.33\,\mathrm{ms}$ windows and
strides.

\begin{table}[t]
\centering
\caption{End-to-end detector latency on an NVIDIA RTX~3090 (ms).}
\label{tab:e2e_latency}
\setlength{\tabcolsep}{3.2pt}
\begin{tabular}{lccccc}
\toprule
Method & Venue & Mean  & Median  & P90  & P95  \\
\midrule
RVT~\cite{rvt}
    & \textcolor{venuegray}{CVPR'23}
    & 29.89 & 28.81 & 37.27 & 41.65 \\
SAST~\cite{sast}
    & \textcolor{venuegray}{CVPR'24}
    & 47.86 & 47.05 & 56.36 & 58.64 \\
S5-ViT~\cite{ssm}
    & \textcolor{venuegray}{CVPR'24}
    & 38.38 & 37.26 & 46.11 & 49.84 \\
SMamba~\cite{smamba}
    & \textcolor{venuegray}{AAAI'25}
    & 45.17 & 43.40 & 58.20 & 59.48 \\
\bottomrule
\end{tabular}
\end{table}

\paragraph{Computational overhead}
Under identical conditions on a single NVIDIA RTX~3090, ChronoFuse
adds only \(0.84\,\mathrm{ms}\) (\(2.84\%\)) of mean end-to-end
latency relative to RVT-B. The fusion pathway adds just
\(0.17\)M parameters (\(0.93\%\)) and a \(1.64\,\mathrm{MiB}\)
feature cache per stream, requires no additional encoder pass, and
keeps the mean runtime below the \(33.33\,\mathrm{ms}\) prediction
horizon.

\paragraph{Baselines}
We compare against four representative event-based detectors spanning recurrent Transformer, sparse Transformer, state-space, and Mamba-based architectures: RVT~\cite{rvt}, SAST~\cite{sast}, S5-ViT~\cite{ssm}, and SMamba~\cite{smamba}. We additionally evaluate causal Constant Velocity and Kalman extrapolation on RVT detections. Kalman provides a strong classical baseline by adaptively estimating object-specific motion while suppressing detection noise; its linear formulation also makes an EKF unnecessary. All extrapolation variants predict detections at $t+\Delta$ using only information available up to $t$.


\subsection{Main Results}
\label{sec:main_results}

\begin{table*}[t]
    \centering
    \caption{
    Detection performance at a fixed $\Delta=33.33$ms latency, chosen as a common horizon representative of measured end-to-end detector runtimes. Existing detector outputs are evaluated at $t+\Delta$; motion baselines extrapolate to the same horizon.
    }
    \label{tab:main_results}

    \setlength{\tabcolsep}{4pt}
    \renewcommand{\arraystretch}{1.08}

    \resizebox{\textwidth}{!}{
    \begin{tabular}{@{}l*{9}{r}@{}}
        \toprule

        &
        \multicolumn{3}{c}{1Mpx}
        &
        \multicolumn{3}{c}{FRED}
        &
        \multicolumn{3}{c}{EV-Flying}
        \\

        \cmidrule(lr){2-4}
        \cmidrule(lr){5-7}
        \cmidrule(lr){8-10}

        Method
        & sAP & sAP$_{50}$ & sAP$_{75}$
        & sAP & sAP$_{50}$ & sAP$_{75}$
        & sAP & sAP$_{50}$ & sAP$_{75}$
        \\
        \midrule

        RVT \textcolor{venuegray}{(CVPR'23)}
        & 43.17 & 73.70 & 42.84
        & 26.88 & 76.22 & 12.25
        & 1.94 & 10.47 & 0.06
        \\

        S5-ViT-B \textcolor{venuegray}{(CVPR'24)}
        & 43.37 & 71.89 & 44.55
        & 25.29 & 72.61 & 10.69
        & 2.01 & 10.71 & 0.07
        \\

        SAST \textcolor{venuegray}{(CVPR'24)}
        & 43.43 & 74.89 & 43.31
        & 27.17 & 76.75 & 12.46
        & 1.86 & 9.75 & 0.07
        \\

        SMamba \textcolor{venuegray}{(AAAI'25)}
        & 43.59 & \textbf{75.44} & 43.00
        & 25.36 & 73.74 & 10.40
        & 2.25 & 11.41 & 0.06
        \\

        \midrule

        RVT + Constant Velocity
        & 44.72 & 73.77 & 45.33
        & 39.75 & 81.38 & \underline{34.40}
        & 3.97 & 14.04 & 1.30
        \\

        RVT + Kalman
        & \underline{45.55} & 74.50 & \underline{46.43}
        & \underline{39.98} & \underline{84.90} & 31.30
        & \underline{7.91} & \underline{21.77} & \underline{3.50}
        \\

        \midrule

        \textbf{ChronoFuse (Ours)}
        & \textbf{46.20} & \underline{74.93} & \textbf{47.60}
        & \textbf{44.99} & \textbf{87.61} & \textbf{40.93}
        & \textbf{20.95} & \textbf{54.66} & \textbf{9.87}
        \\

        \bottomrule
    \end{tabular}
    }
\end{table*}

\begin{figure*}[t]
    \centering
    \includegraphics[width=\textwidth]
    {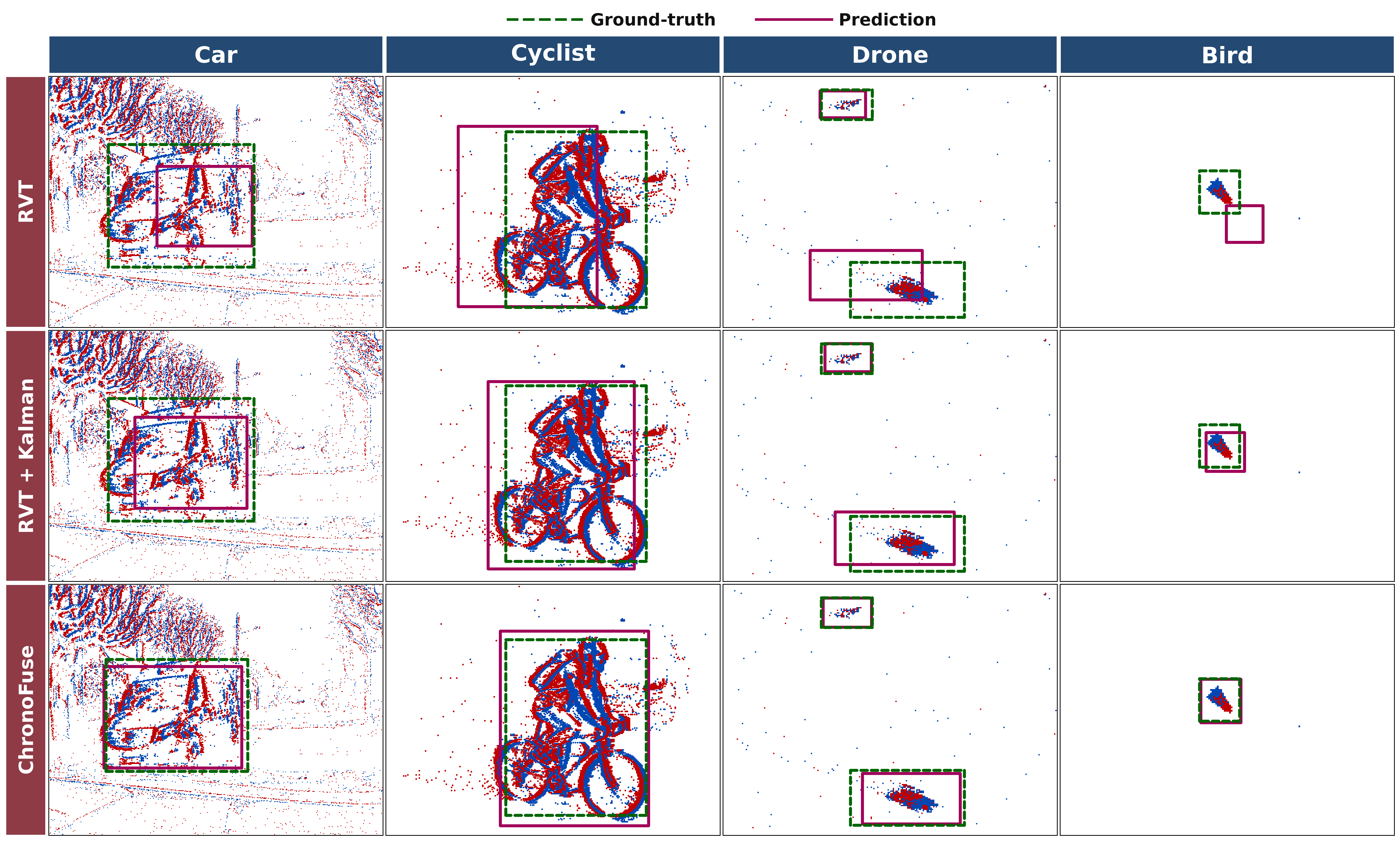}
    \caption{Qualitative comparison across representative motion regimes.
Green dashed boxes denote ground truth and solid magenta boxes denote predictions.
Compared with RVT and RVT + Kalman extrapolation at the same availability
horizon, ChronoFuse maintains tighter spatial alignment across motion regimes,
including both the near-static (upper) and the fast-moving drone (lower).}
    \label{fig:qualitative}
\end{figure*}

\paragraph{1Mpx}
A $33.33\,\mathrm{ms}$ delay reduces RVT from $47.39$ to $43.17$
sAP, a $4.22$-point loss caused by temporal misalignment. Constant
Velocity and Kalman recover only part of this loss, reaching $44.72$
and $45.55$ sAP, respectively. ChronoFuse raises performance to
$46.20$ sAP, outperforming these baselines by $1.48$ and $0.65$
points. It gains $3.03$ sAP over RVT at prediction availability,
recovers $71.8\%$ of the latency-induced loss, and comes within
$1.19$ points of the zero-delay reference.

ChronoFuse particularly improves precise localization:
$\mathrm{sAP}_{50}$ rises from $73.70$ to $74.93$, whereas
$\mathrm{sAP}_{75}$ increases from $42.84$ to $47.60$. The larger
$4.76$-point gain at the stricter threshold indicates improved spatial
alignment. The gains are also consistent across pedestrians,
two-wheelers, and cars, with class AP improvements of $3.46$, $3.18$,
and $2.45$ points, respectively.

We further rank matched objects by box-normalized image-plane
displacement and partition them into three equal-cardinality motion
groups. The RVT loss caused by latency increases from $0.53$ sAP for
low motion to $2.23$ for medium motion and $10.06$ for high motion.
For high-motion objects, the corresponding $\mathrm{sAP}_{75}$ loss
reaches $15.44$ points. ChronoFuse recovers $1.83$ sAP in the
medium-motion group and $7.76$ sAP in the high-motion group, restoring
$77\%$ of the high-motion loss. This shows that its benefit grows with
temporal displacement, where latency compensation is most critical. Moreover, future-target
supervision transfers beyond RVT: applying FT-only supervision to
SMamba improves availability-time $\mathrm{sAP}$ from $43.59$ to
$45.95$ ($+2.36$).

\paragraph{FRED} FRED exhibits a severe latency penalty owing to the rapid image-plane motion of small drones. RVT achieves \(46.82\) sAP under observation-time evaluation, but only \(26.88\) sAP at the \(33.33\,\mathrm{ms}\) availability horizon, a \(19.94\)-point reduction. Constant Velocity and Kalman recover to \(39.75\) and \(39.98\) sAP, restoring \(64.5\%\) and \(65.7\%\) of this loss, respectively. ChronoFuse reaches \(44.99\) sAP, improving upon RVT at prediction availability by \(18.11\) points, recovering \(90.8\%\) of the lost accuracy, and remaining within \(1.83\) points of the observation-time reference. The improvement is particularly pronounced under strict localization: ChronoFuse gains \(11.39\) points at \(\mathrm{sAP}_{50}\) and \(28.68\) points at \(\mathrm{sAP}_{75}\), raising \(\mathrm{sAP}_{75}\) from \(12.25\) to \(40.93\). This disproportionate gain at the stricter overlap threshold demonstrates its ability to correct detections that would otherwise become spatially stale during computation.

\paragraph{EV-Flying}
EV-Flying provides the most demanding stress test, with tiny birds and
insects moving by a large fraction of their box size within a single
detection interval. Against a same-time RVT reference of $38.41$ sAP,
existing event detectors fall to at most $2.25$ sAP when evaluated at
prediction availability, whereas ChronoFuse reaches $20.95$ sAP---$9.3\times$
the strongest existing-detector result. This regime exposes the failure mode
most directly: small temporal displacements can destroy IoU when the target
itself occupies only a few pixels. Fig.~\ref{fig:qualitative} illustrates
this progression across road users, drones, and fast biological targets.

\begin{figure*}[t]
\centering

\begin{minipage}[b]{0.345\textwidth}
    \centering
    \includegraphics[width=\linewidth]{
        \detokenize{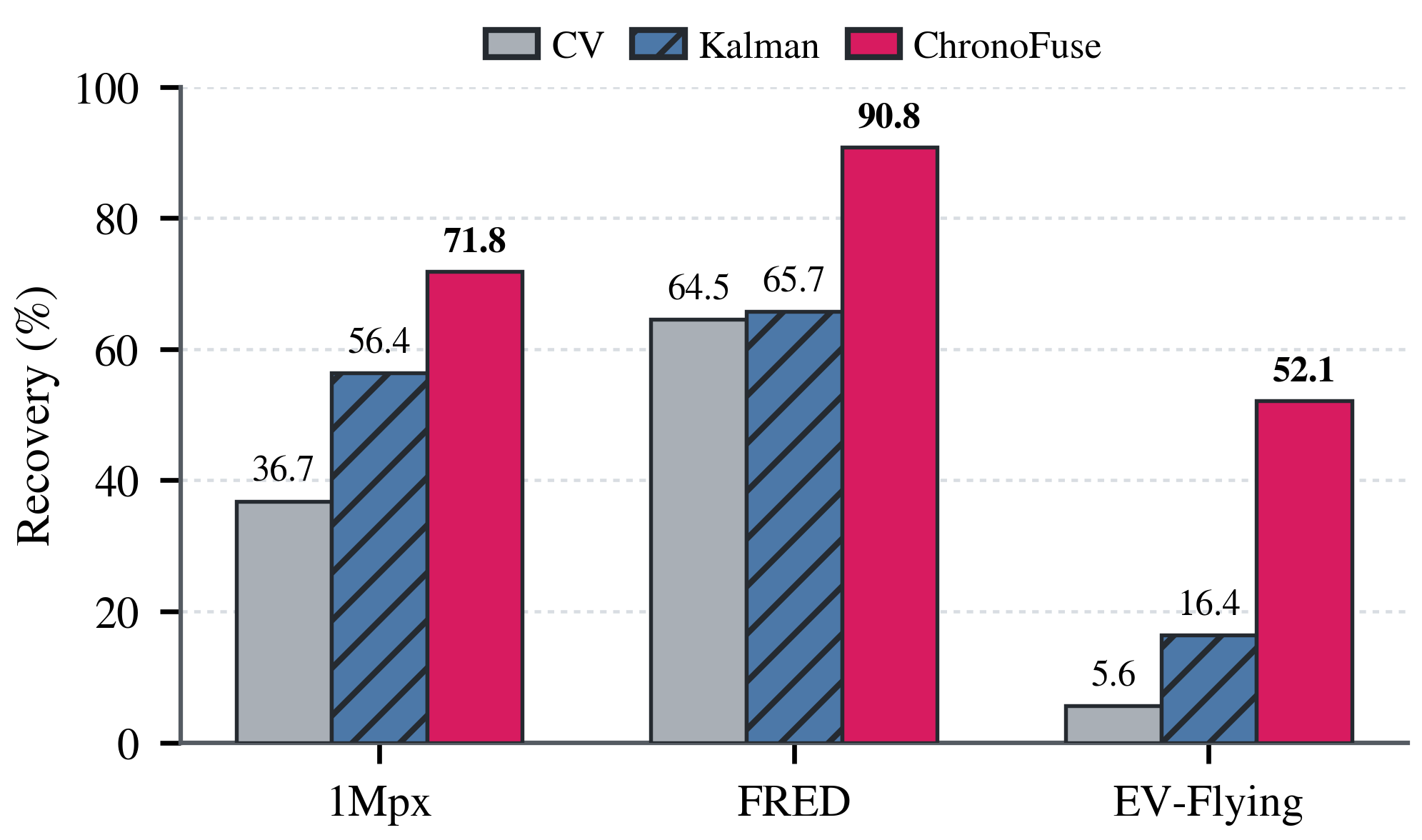}}
    \par\vspace{1mm}
    {\footnotesize\textbf{(a)} Cross-dataset recovery}
\end{minipage}
\hfill
\begin{minipage}[b]{0.345\textwidth}
    \centering
    \includegraphics[width=\linewidth]{
        \detokenize{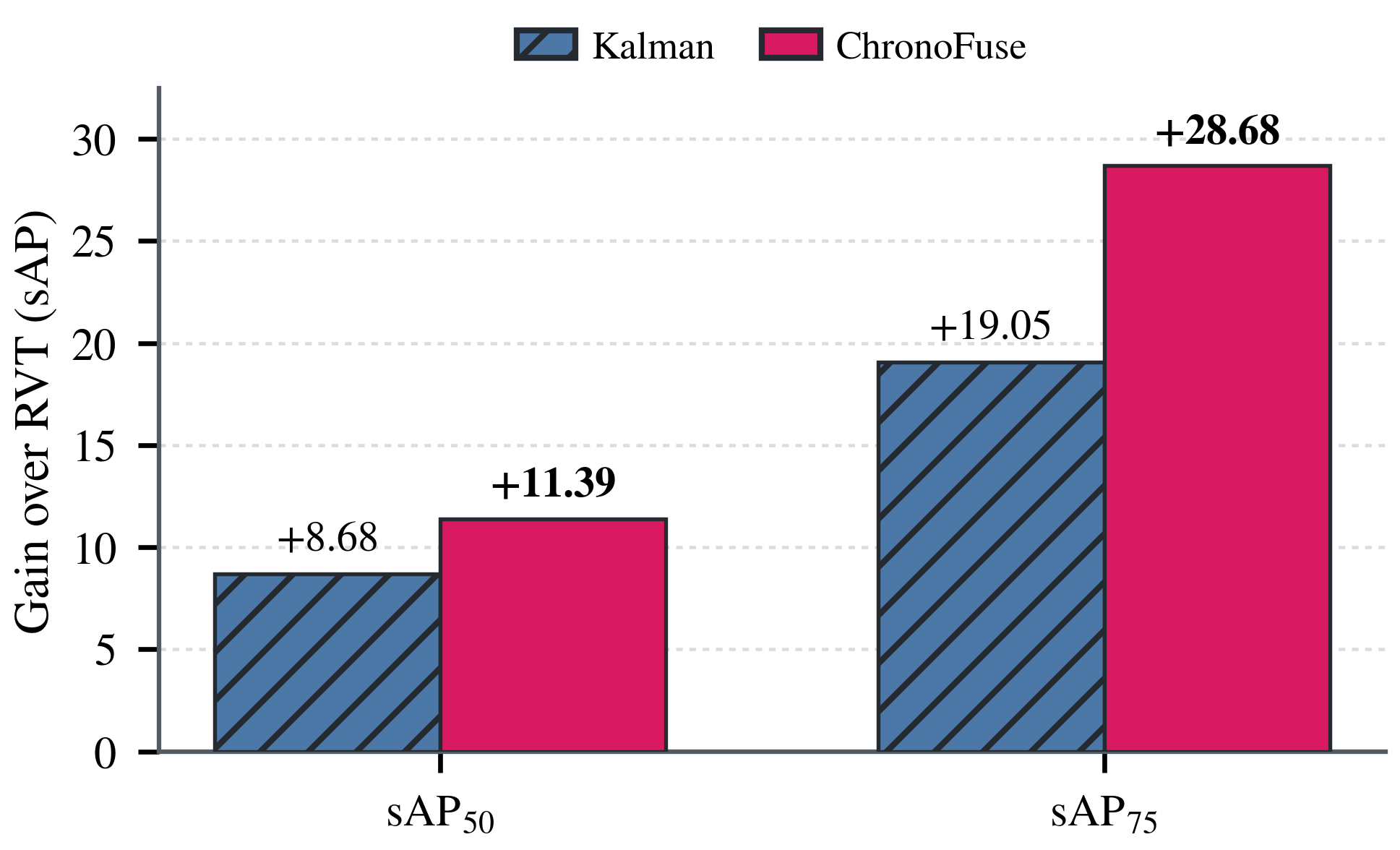}}
    \par\vspace{1mm}
    {\footnotesize\textbf{(b)} FRED localization recovery}
\end{minipage}
\hfill
\begin{minipage}[b]{0.27\textwidth}
    \centering
    \includegraphics[width=\linewidth]{
        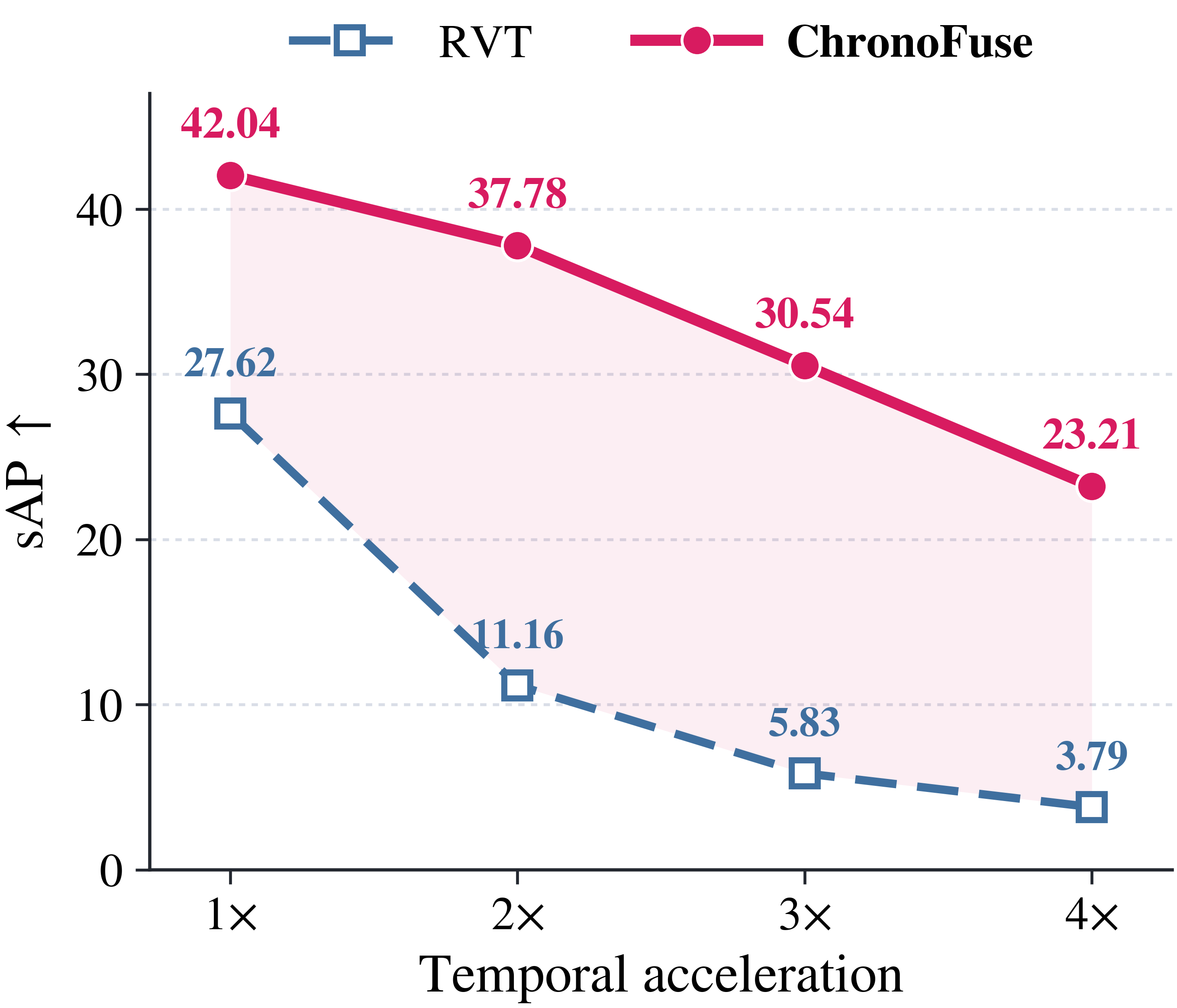}
    \par\vspace{1mm}
    {\footnotesize\textbf{(c)} Motion-speed robustness}
\end{minipage}

\caption{\textbf{Complementary analysis of latency compensation.}
(a) Across 1Mpx, FRED, and EV-Flying, ChronoFuse consistently recovers
a larger fraction of the latency-induced accuracy loss than model-based
motion extrapolation.
(b) On FRED, ChronoFuse produces substantially larger gains at the
stricter overlap threshold, indicating improved localization of objects
displaced during computation.
(c) Under mixed-speed training, ChronoFuse remains substantially more
robust than RVT as motion is accelerated from $1\times$ to $4\times$.}
\label{fig:complementary_analysis}

\end{figure*}

Overall, the results show a consistent trend: as target motion becomes
more severe, stale detection degrades more sharply, while predictive
latency compensation becomes increasingly valuable, as summarized in
Fig.~\ref{fig:complementary_analysis}(a,b).

\subsection{Comparison with RGB Streaming Detection}

We compare ChronoFuse against StreamYOLO~\cite{yang2022real}, a strong RGB
streaming detector, on the corresponding synchronized FRED sequences,
using events for ChronoFuse and RGB frames for StreamYOLO under the same
streaming evaluation protocol. Because the systems use different sensing
modalities and architectures, we present this comparison as a
matched-timing cross-modal reference rather than a controlled modality
ablation.

StreamYOLO achieves $22.76$ sAP ($59.6$ $\mathrm{sAP}_{50}$,
$11.6$ $\mathrm{sAP}_{75}$), whereas ChronoFuse reaches $44.88$ sAP
($87.59$ $\mathrm{sAP}_{50}$, $40.72$ $\mathrm{sAP}_{75}$), yielding
gains of $22.12$ sAP and $29.12$ $\mathrm{sAP}_{75}$. The larger margin
at the stricter overlap threshold indicates substantially tighter
availability-time localization under rapid drone motion.

\subsection{Motion-speed robustness}
We progressively accelerate FRED from $1\times$ to $4\times$ while
keeping the availability horizon fixed
(Fig.~\ref{fig:complementary_analysis}(c) and
Table~\ref{tab:speed_robustness}). With training only at $1\times$,
ChronoFuse substantially outperforms RVT and SMamba at all unseen
speeds, although performance degrades under the largest speed shift.
Mixed-speed training raises ChronoFuse's Avg. sAP from $21.01$ to
$33.39$, versus $12.10$ for RVT and $11.82$ for SMamba.
ChronoFuse retains $30.54$ and $23.21$ sAP at $3\times$ and $4\times$,
respectively, with only a modest reduction in nominal-speed accuracy.

\begin{table}[t]
\centering
\caption{Motion-speed robustness on FRED at $\Delta=33.33$ms.
Mix denotes joint $1\times$--$4\times$ training; Avg. sAP averages
all four speeds.}
\label{tab:speed_robustness}

\setlength{\tabcolsep}{3pt}
\renewcommand{\arraystretch}{1.05}

\resizebox{\columnwidth}{!}{
\begin{tabular}{@{}l*{5}{r}@{}}
    \toprule
    Method
    & $1\times$ & $2\times$ & $3\times$ & $4\times$ & Avg.sAP \\
    \midrule

    RVT$_{1\times}$
    & 26.86 & 8.97 & 4.41 & 2.78 & 10.75 \\

    SMamba$_{1\times}$
    & 25.30 & 9.09 & 4.90 & 3.24 & 10.63 \\

    \textbf{ChronoFuse}$_{1\times}$
    & \textbf{44.88} & \textbf{23.50} & \textbf{10.08}
    & \textbf{5.57} & \textbf{21.01} \\

    \midrule

    RVT$_{\mathrm{Mix}}$
    & 27.62 & 11.16 & 5.83 & 3.79 & 12.10 \\

    SMamba$_{\mathrm{Mix}}$
    & 26.62 & 11.07 & 5.91 & 3.68 & 11.82 \\

    \textbf{ChronoFuse}$_{\mathrm{Mix}}$
    & \textbf{42.04} & \textbf{37.78} & \textbf{30.54}
    & \textbf{23.21} & \textbf{33.39} \\

    \bottomrule
\end{tabular}
}

\end{table}

\subsection{Ablation Study}
\label{sec:ablations}

We ablate ChronoFuse on 1Mpx and EV-Flying under the same future-target
protocol, isolating the two central design choices in
Fig.~\ref{fig:chronofuse_architecture}. FT-only bypasses the Cross-Time
Fusion Module while retaining future supervision, separating the
supervision shift from temporal fusion. The no-residual variant retains
past--present fusion but removes the current-feature identity path,
testing the value of preserving the complete current representation.

ChronoFuse performs best on both datasets. On 1Mpx, removing fusion and
the residual path lowers sAP from $46.20$ to $45.75$ and $44.97$,
respectively. On EV-Flying, the corresponding scores are $16.54$ and
$18.82$, compared with $20.95$ for ChronoFuse, confirming that both
components remain beneficial under rapid motion
(See Table~\ref{tab:ablation}).

\begin{table}[t]
    \centering
    \caption{
        Ablation results (sAP) on 1Mpx and EV-Flying
        ($\Delta=33.33\,\mathrm{ms}$).
    }
    \label{tab:ablation}
    \small
    \setlength{\tabcolsep}{7pt}
    \renewcommand{\arraystretch}{1.08}

    \begin{tabular}{@{}lrr@{}}
        \toprule
        Variant & 1Mpx & EV-Flying \\
        \midrule

        FT-only (w/o fusion)
        & 45.75 & 16.54 \\

        w/o residual connection
        & 44.97 & 18.82 \\

        \midrule

        \textbf{ChronoFuse}
        & \textbf{46.20}
        & \textbf{20.95} \\

        \bottomrule
    \end{tabular}
\end{table}

\section{Conclusion}

We showed that computation latency is not merely a systems-level
detail in event-based robotic perception: even delays of a few tens
of milliseconds can leave detections spatially stale when they
become available, particularly under fast motion. We introduced
\textbf{ChronoFuse}, a causal cross-time fusion framework that
predicts object states directly at availability time using only
observed events, rather than extrapolating already-stale detections.
Across driving, drone, and bird/insect scenarios, ChronoFuse
substantially recovers accuracy lost to latency, including
approximately $71\%$ and $91\%$ of RVT's latency-induced accuracy
loss on 1Mpx and FRED, respectively. These gains extend to
EV-Flying, where rapid bird and insect motion causes the performance of standard event detectors to degrade sharply. Synthetic high-speed stress tests
further expose the vulnerability of observation-time detection
as apparent motion increases. On synchronized FRED data,
ChronoFuse also achieves nearly twice the streaming accuracy
of RGB StreamYOLO under the same evaluation protocol. By aligning detections with the world state at prediction availability
rather than at observation time, ChronoFuse provides more timely object
state estimates for downstream robotic decision-making, including
navigation, interception, and multi-robot coordination. Future work can
extend the current fixed-horizon formulation to latency-conditioned models
that adapt their predictions to changing computational loads and system
delays. Taken together, these results establish availability-time prediction
as an effective paradigm for latency-aware event-based robotic perception,
substantially recovering localization accuracy precisely in the high-speed
regimes where stale detections are most damaging.

\section*{Acknowledgment}
The authors thank the anonymous reviewers for constructive feedback. The computational work involved in this research was partially supported by NUS IT’s Research Computing group under grant NUSREC-HPC-00001. OpenAI Codex was used for coding assistance during implementation and
experimental scripting. OpenAI ChatGPT was used for conceptual
brainstorming and refinement of diagrams in the Introduction and Method
sections. All resulting material was reviewed and validated by the authors.

\bibliographystyle{IEEEtran}
\bibliography{bibilo}

\end{document}